\documentclass{article}
\usepackage{spconf,amsmath,graphicx}

\usepackage{enumitem}
\usepackage[table]{xcolor}
\usepackage{enumitem}
\usepackage{longtable}
\usepackage{subcaption}
\usepackage{amsmath}
\usepackage{graphicx} 
\usepackage{booktabs}
 \usepackage{multirow}

\setlist{nosep, leftmargin=14pt}

\usepackage{mwe} 

\title{Reliable Brain Tumor Segmentation Based on Spiking Neural Networks with Efficient Training}
\name{Aurora Pia Ghiardelli, Guangzhi Tang, Tao Sun \sthanks{Corresponding author: Tao Sun (tao.sun@maastrichtuniversity.nl)} \thanks{This work used the Dutch national e-infrastructure with the support of the SURF Cooperative using grant no. EINF-10787.}}
\address{Department of Advanced Computing Sciences, Maastricht University, Maastricht, The Netherlands}

\usepackage{tikz}
\usepackage{textcomp}
\usepackage{hyperref}
\usepackage{lipsum}

\newcommand\copyrighttext{%
  \footnotesize \textcopyright 2026 IEEE. Personal use of this material is permitted.
  Permission from IEEE must be obtained for all other uses, in any current or future
  media, including reprinting/republishing this material for advertising or promotional
  purposes, creating new collective works, for resale or redistribution to servers or
  lists, or reuse of any copyrighted component of this work in other works.}
\renewcommand\copyrightnotice{%
\begin{tikzpicture}[remember picture,overlay]
\node[anchor=south,yshift=10pt] at (current page.south) {\fbox{\parbox{\dimexpr\textwidth-\fboxsep-\fboxrule\relax}{\copyrighttext}}};
\end{tikzpicture}%
}

\begin{document}
%
\maketitle
\copyrightnotice
\begin{abstract}
We propose a reliable and energy-efficient framework for 3D brain tumor segmentation using spiking neural networks (SNNs). A multi-view ensemble of sagittal, coronal, and axial SNN models provides voxel-wise uncertainty estimation and enhances segmentation robustness. To address the high computational cost in training SNN models for semantic image segmentation, we employ Forward Propagation Through Time (FPTT), which maintains temporal learning efficiency with significantly reduced computational cost. Experiments on the Multimodal Brain Tumor Segmentation Challenges (BraTS 2017 and BraTS 2023) demonstrate competitive accuracy, well-calibrated uncertainty, and an 87\% reduction in FLOPs, underscoring the potential of SNNs for reliable, low-power medical IoT and Point-of-Care systems.
\end{abstract}
\section{Introduction}
\label{sec:intro}

Accurate and reliable brain tumor segmentation is crucial for clinical diagnosis, treatment planning, and prognosis ~\cite{li2024brain}. Therefore, segmentation models must not only deliver precise delineations but also quantify their predictive confidence, allowing clinicians to assess the trustworthiness of results. Such uncertainty-aware modeling is essential for the safe and interpretable integration of AI systems into clinical practice~\cite{gal2016uncertainty}.

Brain tumor datasets typically consist of 3D multi-modal MRI volumes, and the segmentation task involves delineating each volume into distinct tumor subregions at the voxel level~\cite{bakas2017advancing}. Although deep neural networks (DNNs) have achieved remarkable performance in 3D brain tumor segmentation, their high computational and energy demands limit their suitability for medical Internet of Things (IoT) and Point-of-Care (PoC) applications, where devices often operate under strict power constraints~\cite{azghadi2020hardware}.

\begin{figure}[pbt]
  \centering
     \includegraphics[width=0.475\textwidth]{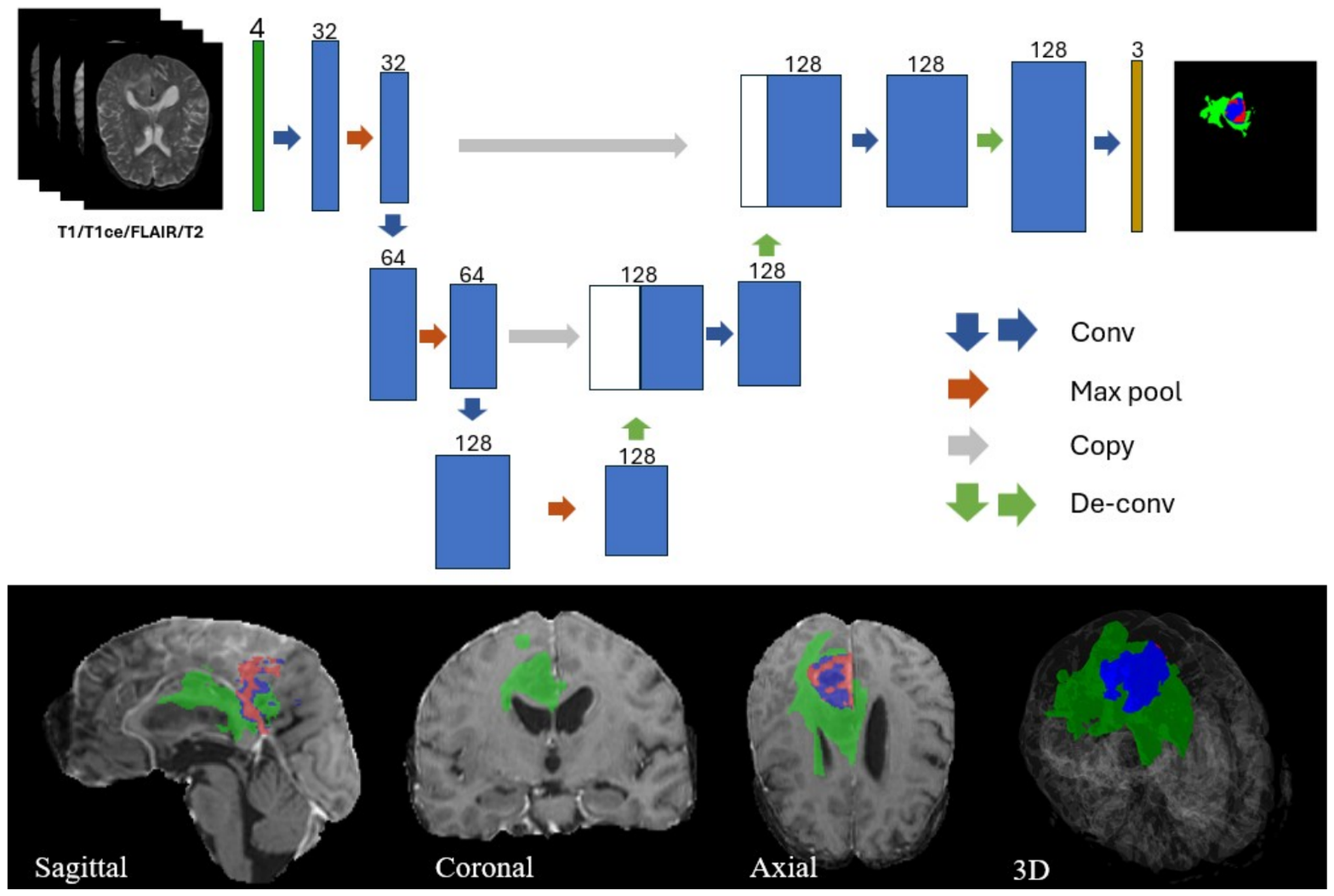}
  \caption{(Top) Architecture of proposed Spiking U-Seg-Net; (Bottom) Tumor segmentation results across three views—sagittal, coronal, and axial—with enhancing tumor in red, necrotic/non-enhancing tumor in blue, and peritumoral edema in green.}
  \label{fig:approaches}
\end{figure}

Spiking neural networks (SNNs), inspired by the event-driven communication of biological neurons, offer an energy-efficient and temporally dynamic alternative. By transmitting information through discrete spike events instead of continuous activations, SNNs perform sparse and stateful computation, significantly reducing energy consumption while maintaining temporal modeling capability~\cite{patel2021spiking}. These characteristics make SNNs particularly suitable for edge deployment on IoT devices and PoC systems, where energy efficiency, low latency, and data privacy are critical.

A 3D brain MRI volume can be examined from three anatomical planes (or \textit{views})—sagittal, coronal, and axial—each offering a distinct 2D perspective of the underlying 3D structure (see Fig.~\ref{fig:approaches}, bottom). SNN-based segmentation models for such 3D voxel data typically treat 2D slices from a given view as sequential inputs, leveraging the inherent temporal processing capability of SNNs. Although recent work has explored uncertainty estimation in SNNs to improve model reliability~\cite{Sun2023Efficient,sun2024aot}, these methods assume a static input repeated across time steps, a setting that is incompatible with the sequential slice-wise inputs required in SNN-based 3D medical image segmentation.

Inspired by the principle of deep ensembles~\cite{lakshminarayanan2017simple}, the gold standard for predictive uncertainty estimation in deep learning~\cite{wilson2020bayesian}, and multi-view ensemble ConvNets \cite{chen2017hippocampus}, we propose a multi-view ensemble strategy. In this approach, an SNN architecture named \textit{Spiking U-Seg-Net} is designed, and three models, one for each view, are trained independently based on this architecture and then combined into an ensemble. This ensemble provides voxel-wise uncertainty estimation that fully leverages the complementary information across anatomical views, thereby enhancing both the reliability and interpretability of the model’s predictions for clinical use~\cite{lakshminarayanan2017simple}.




Training SNNs for brain tumor segmentation remains challenging. Early SNN training approach based on DNN-to-SNN conversion \cite{rueckauer2017conversion} simplified training by mapping pretrained activations to spike rates, but it is limited by architectural constraints and quantization errors. More recent direct training methods employ backpropagation through time (BPTT) with surrogate gradients \cite{surrogate_gradient} to address the non-differentiable spiking mechanism; however, these methods entail substantial, sometimes prohibitive, computational and memory burdens \cite{li2025efficient}.

To overcome these limitations, we adopt \textit{Forward Propagation Through Time} (FPTT)~\cite{pmlr-v139-kag21a,yin2023accurate}, a training algorithm that computes temporal gradients in a forward manner without explicit backtracking. FPTT enables efficient and scalable optimization of temporally dynamic SNNs while maintaining accurate temporal credit assignment, making it well-suited for training multi-layer medical imaging models under limited computational resources.

In summary, our main contributions are:
(i) an energy-efficient SNN-UNet architecture trained with FPTT for brain tumor segmentation;
(ii) a multi-view ensemble inference mechanism that improves both robustness and uncertainty estimation; and
(iii) a comprehensive evaluation on the Multimodal Brain Tumor Segmentation Challenges (BraTS 2017~\cite{bakas2017advancing} and BraTS 2023~\cite{li2024brain}), demonstrating competitive accuracy, improved reliability, and high energy efficiency suitable for medical IoT and point-of-care (PoC) systems.

\section{Background}
\subsection{Uncertainty Estimation and Deep Ensemble}
In machine learning models, uncertainty is typically represented through probabilities \cite{ghahramani2015probabilistic}. In image semantic segmentation, high-quality uncertainty estimation ensures that predicted probability for each pixel accurately reflects the true likelihood of the prediction being correct. Uncertainty comprises two components: aleatoric uncertainty, arising from inherent data noise, and epistemic uncertainty, caused by limited data and model stochasticity. Uncertainty estimation methods can be classified into deterministic approaches, such as deep ensembles~\cite{lakshminarayanan2017simple}, and Bayesian methods, such as Monte Carlo dropout~\cite{gal2016uncertainty}. Deep ensembles are widely regarded as the gold standard, offering reliable and well-calibrated predictive confidence~\cite{wilson2020bayesian}. Deep ensembles estimate epistemic uncertainty by training multiple DNNs independently and averaging their predictions, where the variance across models reflects the model’s confidence.

\vspace{-0.5em}

\subsection{Spiking Neural Networks (SNNs)}
SNNs resemble DNNs structurally but differ in computation across discrete time steps. At each time step $t = 0, 1, \ldots, T$, the network processes the input $x_t$ and produces a prediction $\hat{y}_t$ to approximate the target $y_t$, leveraging \textit{stateful} neurons that emit binary spikes. The membrane potential $u_t$, representing the neuron's internal state, evolves from presynaptic spikes and its previous state $u_{t-1}$, and the neuron fires ($s_t = 1$) when $u_t$ exceeds the threshold $\theta$. This temporal, event-driven mechanism yields sparse, energy-efficient, and low-latency computation.
\vspace{-0.5em}
\subsubsection{Leaky Integrate-and-Fire (LIF) Neurons}
Among spiking neuron models, the Leaky Integrate-and-Fire (LIF) model~\cite{Gerstner2002-wd} is widely adopted for its balance between biological plausibility and computational simplicity. Its membrane potential dynamics are governed by $\tau \frac{du}{dt} = -u + RI$,
where $I$ is the input current, $R$ the membrane resistance, and $\tau$ the membrane time constant.
In discrete form, this can be expressed as $u_t = \lambda u_{t-1} + \sum_j ws_t - \theta s_{t-1}$, where $\lambda < 1$ is the leak constant, $w$ the synaptic weight, and $s_t \in \{0,1\}$ the binary spike output.
\vspace{-0.5em}
\subsubsection{Training of SNNs} 
Two main approaches exist for training deep SNNs: DNN-to-SNN conversion \cite{rueckauer2017conversion} and direct training with surrogate gradients \cite{surrogate_gradient}. DNN-to-SNN conversion, used in recent image segmentation studies~\cite{biswas2024halsie}, is limited by the need for large time steps, sensitivity to architectural and normalization constraints, and degradation from quantization errors. For segmentation, it further neglects the interaction between temporal dynamics and spatial features. In contrast, direct training employs surrogate gradients~\cite{surrogate_gradient} to enable BPTT. While better suited for temporal data, it remains computationally demanding and unstable, and often underperforms DNNs on complex tasks such as semantic segmentation. As a result, recent work increasingly adopts hybrid DNN+SNN models to mitigate these issues, combining DNN spatial encoding with the temporal efficiency of SNNs \cite{li2025efficient}.

\section{Method}

\subsection{Network Architecture}
The architecture of Spiking U-Seg-Net (see Fig. \ref{fig:approaches}, top) follows the encoder–decoder structure of the conventional U-Net, where the encoder progressively abstracts spatial features through convolution and pooling operations, and the decoder reconstructs fine-grained segmentation maps via deconvolution and skip connections that fuse multi-scale feature representations. Each input slice consists of four MRI modalities, namely T1-weighted (T1), T2-weighted (T2), T1-weighted with gadolinium contrast enhancement (T1-Gd), and Fluid Attenuated Inversion Recovery (FLAIR), stacked as separate channels. Padding is applied in every convolutional layer to preserve spatial dimensions. Unlike the original U-Net, which uses two convolutions per block, we employ a single convolution per block to reduce the number of learnable parameters and computational cost. Group Normalization is applied after each convolution to improve training stability with small batch sizes, and dropout is introduced at multiple layers to enhance generalization. After dropout, Parametric LIF (PLIF) neurons~\cite{fang2021incorporating} are applied, where the membrane time constant $\tau$ is learnable and shared across neurons within each layer. The surrogate gradient function is defined as $\sigma(x) = \frac{1}{\pi} \arctan(\pi x) + \frac{1}{2}$. The final convolutional readout integrator layer employs $3\times3$ kernels to produce three output feature maps. These maps correspond to predictions of the enhancing tumor (ET, red), the tumor core (TC), which includes ET and necrotic or non-enhancing tumor regions (NCR/NET, blue), and the whole tumor (WT), encompassing all subregions including peritumoral edema (PE, green) (see Fig.~\ref{fig:approaches}).
\vspace{-0.75em}
\subsection{SNN Inference and Memory-efficient FPTT Training}

In Spiking U-Seg-Net, we leverage the temporal dynamics of spiking neurons to propagate information along the slicing axis of the 3D volume. Each time step corresponds to a single slice of the MRI volume. BPTT requires memory that scales linearly with sequence length. Consequently, given the large number of slices, training SNNs for medical image segmentation with BPTT is prone to vanishing or exploding gradients and often becomes computationally prohibitive~\cite{li2025efficient}.

In this work, we adopt FPTT, which enables efficient optimization of temporal models without backtracking through all past states. Originally proposed for recurrent neural networks~\cite{pmlr-v139-kag21a} and later extended to spiking neural networks~\cite{yin2023accurate}, FPTT updates network parameters at each time step $t$ using an instantaneous risk function $\mathcal{L}_t = \mathcal{L}(y_t, \hat{y}_t) + R(w_t)$, where $\mathcal{L}(y_t, \hat{y}_t)$ is the loss between the ground truth $y_t$ and the prediction $\hat{y}_t$, and $R(w_t)$ is a regularization term that stabilizes the temporal updates of the weights $w_t$. $R(w_t)$ relies on a running average of past weights $\overline{w}_{t}$ and is defined as 
\[
R(w_t) = \frac{\alpha}{2}\left\| w_t - \overline{w}_{t} - \frac{1}{2\alpha}\nabla l_{t-1}(w_t) \right\|^2, 
\]
with parameter updates given by  
\[
\begin{split}
\nabla l_{t}(w_{t+1}) &= \nabla l_{t-1}(w_t) - \alpha (w_t - \overline{w}_{t-1}), \\
\overline{w}_{t+1} &= \tfrac{1}{2} (\overline{w}_{t} + w_{t+1}) - \tfrac{1}{2\alpha} \nabla l_{t}(w_{t+1}).
\end{split}
\]
Here, $\alpha$ controls the strength of the regularization in $\mathcal{L}_t$. This formulation ensures smooth temporal updates and convergence to a stationary solution of the standard SNN objective~\cite{yin2023accurate}.

\vspace{-0.5em}

\subsection{Multi-View Ensemble}
As mentioned previously, a 3D brain MRI volume can be viewed from three orthogonal views: sagittal, coronal, and axial. To enhance uncertainty estimation in brain tumor segmentation, we adopt a multi-view ensemble strategy inspired by the principle of deep ensembles~\cite{lakshminarayanan2017simple} and multi-view ensemble ConvNets \cite{chen2017hippocampus}, which aggregates predictions from the three anatomical views. Each view is modeled independently using a dedicated Spiking U-Seg-Net trained on its corresponding 2D slice sequences. During inference, the probability maps produced by the three models are spatially aligned and averaged voxel-wise to generate the final segmentation.

\vspace{-0.5em}
\subsection{Loss Function} \label{sec:loss}
For training, we employ a hybrid loss that combines the binary cross-entropy loss $L_{BCE}$ and the Dice loss $L_{Dice}$. The Dice loss is derived from the Dice ratio, which quantifies spatial overlap between predictions and ground truth and is defined as $Dice = \frac{2|P \cap G|}{|P| + |G|}$,
where $P$ and $G$ denote the predicted and ground-truth voxel sets, respectively. 
The Dice loss converts the Dice coefficient into a differentiable form for gradient-based optimization $L_{Dice} = 1 - \frac{2\sum_i s_i r_i + \epsilon}{\sum_i s_i + \sum_i r_i + \epsilon}$, where $s_i$ and $r_i$ denote the predicted probability and ground-truth label at pixel $i$, respectively, and $\epsilon = 10^{-5}$ ensures numerical stability. It improves sensitivity to small or imbalanced tumor regions. The final training objective combines both terms equally $L_{total} = 0.5 \times L_{BCE} + 0.5 \times L_{Dice}$.

\section{Experiments and Results}

\subsection{BraTS Dataset}
The data used in this study comes from BraTS 2017~\cite{bakas2017advancing} and BraTS 2023~\cite{li2024brain}, which both provide multi-contrast MRI scans of glioma patients together with expert-refined tumor annotations. Each subject includes four MRI modalities, T1, T2, T1-Gd, and FLAIR, and a multi-class tumor mask delineating enhancing tumor (ET), necrotic/non-enhancing tumor core (NCR/NET), and peritumoral edema (PE). The two datasets contain 210 and 1,251 subjects, respectively.

All MRI volumes have a standardized spatial resolution of $240 \times 240 \times 155$ voxels with 1 mm isotropic spacing through co-registration, resampling, and skull stripping. To remove non-informative peripheral regions, the central region of each volume was cropped to $160 \times 192 \times 152$ voxels. From the cropped data, 2D slices were extracted along the coronal and sagittal views, resulting in 192 coronal and 160 sagittal slices per subject. Each modality and anatomical view was then independently normalized to the range [0, 1] using min–max scaling applied across the entire volume.  
\vspace{-0.75em}
\subsection{Evaluation}
We evaluate model performance using two complementary measures: the 3D Dice ratio for segmentation accuracy and the Negative Log-Likelihood (NLL) for reliability and calibration assessment. The definition of Dice ratio is introduced in subsection \ref{sec:loss}.

To assess the calibration of predicted probability maps, we use the NLL ~\cite{Gneiting2007Proper} , which quantifies the agreement between all the predicted probabilities and observed labels~\cite{Sun2023Efficient}.  
A lower NLL value indicates better-calibrated and more reliable predictions.

We adopt U-Seg-Net and U-Seg-Net+CLSTM from~\cite{chen2019deep} as baselines, as our model follows a similar architecture but replaces the DNN components with spiking neural units. U-Seg-Net performs per-slice 2D segmentation and reconstructs the 3D volume by stacking predictions, whereas U-Seg-Net+CLSTM additionally captures inter-slice context via convolutional LSTMs. Note that both baselines are trained in a multi-view setting, and their view-specific features are fused using the convolutional network proposed in~\cite{chen2017hippocampus}, which is designed to improve segmentation accuracy rather than uncertainty estimation.

\vspace{-0.5em}
\subsection{Training and Testing}

To conduct our experiments, we adopt a 5-fold cross-validation scheme. For BraTS 2017, each fold uses 168 subjects for training and 42 subjects for testing. For BraTS 2023, which contains 1,251 cases in total, the five folds consist of 251, 250, 250, 250, and 250 subjects, respectively. For each fold, three independent Spiking U-Seg-Net models are trained, one per anatomical view. Segmentation is first performed independently for the sagittal, coronal, and axial views, after which the resulting probability maps are fused by voxel-wise averaging to obtain the final ensemble output. All models are implemented in the PyTorch framework and trained with a batch size of 8 using the Adam optimizer with a learning rate of 0.001 and a weight decay of $1\times10^{-5}$. The learning rate is adjusted using a ReduceLROnPlateau scheduler, and gradient clipping with a maximum norm of 0.3 is applied to ensure training stability. Through a grid search, we identified $\alpha = 0.1$ as the optimal parameter for FPTT. 


\begin{table}[t]
\centering
\small
\caption{Dice ratio (\%) of the ensemble SNN and DNN baselines on BraTS17 and BraTS23 (mean $\pm$ std across folds).}
\label{tab:dice}
\scalebox{0.7}{
\begin{tabular}{l lccc}
\toprule
Dataset & Method & ET & TC & WT \\
\midrule
\multirow{3}{*}{BraTS17}
  & Spiking U-Seg-Net \textbf{(ours)} & 78.39$\pm$0.43 & 82.70$\pm$1.09 & 84.59$\pm$2.04 \\
  & U-Seg-Net \cite{chen2019deep} & 78.77$\pm$1.99 & 83.53$\pm$2.23 & 83.89$\pm$2.57 \\
  & U-Seg-Net+CLSTM \cite{chen2019deep} & 79.03$\pm$2.29 & 84.12$\pm$2.60 & 85.10$\pm$2.31 \\
\midrule
\multirow{2}{*}{BraTS23}
  & Spiking U-Seg-Net \textbf{(ours)} & 82.72$\pm$1.14 & 
                               85.87$\pm$1.39 & 
                               89.45$\pm$0.62 \\
  & U-Seg-Net (our run) & 84.10$\pm$0.92 & 87.54$\pm$1.13 & 89.08$\pm$0.82 \\
\bottomrule
\end{tabular}
}
\label{tab:isbi_brats_results}
\end{table}

\vspace{-0.75em}

\subsection{Results}
Table \ref{tab:isbi_brats_results} summarizes the segmentation performance of the proposed Spiking U-Seg-Net and the DNN baselines on the BraTS17 and BraTS23 datasets. On both datasets, the ensemble Spiking U-Seg-Net consistently achieves strong Dice performance across all tumor subregions, with noticeably higher scores on BraTS23 due to its larger size and greater anatomical diversity. Overall, our ensemble results approach both baselines in \cite{chen2019deep} on BraTS17, and achieve superior performance on the whole tumor (WT) region on BraTS23..

The uncertainty estimation results in Table \ref{tab:nll-final} further highlight the benefit of the multi-view ensemble. On BraTS17, the best single-view NLL is 0.0201 (axial), while the ensemble reduces this to 0.0113, corresponding to a relative improvement of approximately 43.8\%. On BraTS23, this improvement (48.3\%) is more significant. This consistent reduction across both datasets indicates that the ensemble not only strengthens segmentation accuracy but also provides significantly better-calibrated uncertainty estimates.

Together, the Dice and NLL results demonstrate that our multi-view ensemble SNN remains robust across datasets of different sizes, delivering both accurate segmentations and reliable probabilistic predictions.

\begin{table}[t]
\centering
\small
\caption{The uncertainty estimaion in NLL for BraTS17 and BraTS23 (mean $\pm$ std across folds).}
\label{tab:nll-final}
\scalebox{0.72}{
\begin{tabular}{lcccc}
\toprule
Dataset & Sagittal & Coronal & Axial & Ensemble \\
\midrule
BraTS17 
& 0.0210 $\pm$ 0.0042 
& 0.0251 $\pm$ 0.0063 
& \textbf{0.0201} $\pm$ 0.0047 
& \textbf{0.0113} $\pm$ 0.0024 \\

BraTS23 
& 0.0150 $\pm$ 0.0008
& \textbf{0.0145} $\pm$ 0.0025
& 0.0155 $\pm$ 0.0015
& \textbf{0.0075} $\pm$ 0.0009 \\
\bottomrule
\end{tabular}}
\end{table}

\subsection{Computational Efficiency} To evaluate computational efficiency of sparse SNN, Table~\ref{tab:flops} reports the number of floating-point operations (FLOPs) required to process a single sample. On average, our SNN achieves an 87\% FLOPs reduction compared to the baseline.


\begin{table}[h]
\centering
\renewcommand{\arraystretch}{1.25}
\small
\footnotesize \resizebox{\columnwidth}{!}{%
\scalebox{0.8}{
\begin{tabular}{lccc|c}
\hline
\textbf{} & \textbf{Sagittal} & \textbf{Coronal} & \textbf{Axial} & \textbf{Ensemble} \\
\hline
\textbf{SNN (ours)}  & $\mathbf{3.15\times10^{11}}$ & $\mathbf{2.97\times10^{11}}$ & $\mathbf{3.38\times10^{11}}$ & $\mathbf{9.5\times10^{11}}$ \\
U-Seg-Net\cite{chen2019deep} & $2.59\times10^{12}$ & $2.16\times10^{12}$ & $2.73\times10^{12}$ & $7.48\times10^{12}$ \\
\hline
\end{tabular}}%
}
\caption{Comparison of computational cost.}
\label{tab:flops}
\end{table}
\vspace{-0.5em}
\section{Conclusion}
The proposed Spiking U-Seg-Net integrates FPTT-based training to efficiently optimize temporal dynamics while reducing computational cost. Through multi-view ensemble inference, it achieves accurate segmentation with well-calibrated predictive uncertainty, offering both energy efficiency and reliability. These results demonstrate the promise of SNNs as a foundation for uncertainty-aware and resource-efficient medical imaging.

\bibliographystyle{IEEEbib}   
\bibliography{references}

\end{document}